# AUTOMATED WORD PREDICTION IN BANGLA LANGUAGE USING STOCHASTIC LANGUAGE MODELS

Md. Masudul Haque [1], Md. Tarek Habib[2] and Md. Mokhlesur Rahman[3]

[1]Dept. of Electrical and Computer Engineering, North South University, Bangladesh
[2]Dept. of Computer Science and Engineering, Daffodil International University, Bangladesh
[3]Dept. of Computer Science and Engineering, Prime University, Bangladesh

*ABSTRACT*

*Word completion and word prediction are two important phenomena in typing that benefit users who type using keyboard or other similar devices. They can have profound impact on the typing of disable people. Our work is based on word prediction on Bangla sentence by using stochastic, i.e. N-gram language model such as unigram, bigram, trigram, deleted Interpolation and backoff models for auto completing a sentence by predicting a correct word in a sentence which saves time and keystrokes of typing and also reduces misspelling. We use large data corpus of Bangla language of different word types to predict correct word with the accuracy as much as possible. We have found promising results. We hope that our work will impact on the baseline for automated Bangla typing.*

*KEYWORDS*

*Word prediction, stochastic model, natural language processing, corpus, N-gram, deleted interpolation, backoff method.*

## 1. INTRODUCTION

Auto complete or word completion works so that the user types the first letter or letters of a word and the program provides one or more higher probable words. If the word he intends to type is included in the list he can select it, for example by using the number of keys. If the word that the user wants is not predicted, the user must type the next letter of the predicted word. At this time, the word choice(s) is altered so that the words provided begin with the same letters as those that have been selected or the word that the user wants appears it is selected. Word prediction technique predicts word by analyzing previous word flow for auto completing a sentence with more accuracy by saving maximum keystroke of any user or student and also reduces misspelling. *N*-gram language model is important technique for word prediction. We use large data corpus for training in *N*-gram language model for predicting correct Bangla word to complete a Bangla sentence with more accuracy.

Word prediction means guessing the next word in a sentence . Word prediction helps disabled people for typing,speed up typing speed by decreasing keystrokes,helps in spelling and error detection and it also helps in speech recognition and hand writing recognition. Auto completion decreases misspelling of word. Word completion and word prediction also helps student to spell any word correctly and to type anything with fewer errors [1].

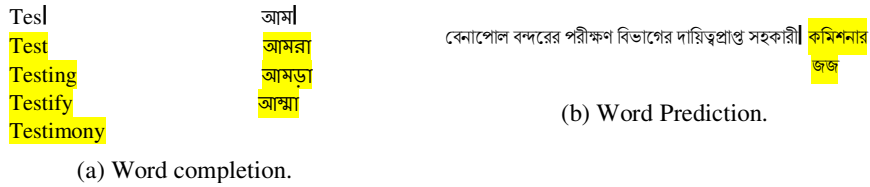

(a) Word completion.  (b) Word Prediction.

Figure 1.Word completion vs. word prediction. Suggested words are highlighted in yellow color.

We survey many techniques to predict upcoming words of a sentence in different languages especially for English Language But there is no satisfactory analysis on Bangla language to predict words in a sentence. So we apply some *N*-gram language model, backoff and deleted interpolation techniques to predict Bangla words in a sentence. Word prediction is very important and complex task in natural language processing (NLP) to predict the correct word to complete a sentence in a very meaningful way.

We use statistical prediction technique such as *N*-gram technique as for example unigram, bigram, trigram, bakeoff propagation, deleted interpolation. We also use large data set of text word in Bangla which is collected from different news paper.

The paper is constructed as follows: related work in section 2, introduction of *N*-gram based word prediction in section 3, methodology in section 4, implementation in section 5, result analysis in section 6 and conclusion in section 7.

## 2. RELATED WORK

In an analysis of predicting sentences [2] researcher developed a sentence completion method based on *N*-gram language models and they derived a *k* best Viterbi beam search decoder for strongly completing a sentence. We also observed use of Artificial Intelligence [3] for word prediction. Here syntactic and semantic analysis is done using the *chart* bottom-up technique for word prediction. Another researcher suggests an approach [4] of word prediction via a Clustered Optimal Binary Search Tree. They suggest using a cluster of computer to build optimal binary search tree which also contain extra link so that bigram and the trigram of the language also presented to achieve optimal performance of word prediction. Here the researcher does a lexical analysis of most probable appeared word in users' text.

In a paper of a learning classification based approach word prediction [5] they present an effective method of word prediction using machine learning and new feature extraction and selection techniques adapted from Mutual Information (MI) and Chi square ($X^2$). Some researchers use *N*-gram language model for word completion in Urdu language [6] and in Hindi language [7] for detecting disambiguation in Hindi word. There are some related works also on Bangla language using *N*-gram language model such as grammar checker of Bangla language [8], checking the correctness of Bangla word [9] and verification of Bangla sentence structure [10].

There are different word prediction tools such as AutoComplete by Microsoft, AutoFill by Google Chrome, TypingAid [11], LetMeType [12] etc. There are some tools for word completion in Bangla language such as Avro [13] . Some other  Bangla software provide word completion features only but theses software do not  give word prediction or sentence completion features. That is why we show word prediction process by using *N*-gram language model to complete a Bangla sentence in our paper.

## 3. *N*-GRAM BASED WORD PREDICTION

*N*-gram language model is a type of probabilistic language model where the approximate matching of next item is very high. Probability is based on counting things or word in most cases. The probability of a word depends on the previous word which is called Markov assumption. Unigram looks single item from a given sequence. Bigram is called first-order Markov model which looks one word into the past and trigram is second-order Markov model which looks two words into the past and quadrigram is third-order Markov model which looks three words into the past and similarly an *N*-gram language model is *N*-1 Markov model which looks *N*-1 words into the past[14]. Thus the general equation for this *N*-gram approximation to the conditional probability of the next word in a sequence, $w_1, w_2, ..., w_n$, is:

$$P(w_n | w_1^{n-1}) \approx P(w_n | w_{n-N+1}^{n-1}) \qquad (1)$$

If *N* = 1, 2, 3 in (1), the model becomes unigram, bigram and trigram language model, respectively, and so on.

For example, using unigram probability of the sentence " I want to eat " is,
P(I want to eat) = P(I) × P(want) × P(to) × P(eat)

Now, using bigram probability of the sentence "I want to eat " is ,
P(I want to eat) = P( I | <start>) × P(want| I) × P(to | want) ×P( eat | to)

Now, using trigram probability of the sentence "I want to eat " is ,
P(coming back to the) = P(I | <s> <s>) × P(want | <s> I) × P(to | I want) × P(eat | want to)

Now if any of these four words are not in the training corpus then the probability of the sentence will be zero for the cause of multiplication. So in this statistical method if we want to consider these words then we need a huge data corpus that must contain all the words of the language. So there may arise the problems like:

- Many entries in corpus are with low frequency
- Probability of a word sequence will be very low or zero

If an entry does not exist in corpus then the probability of the sentence will become zero because of multiplication.

To solve the problem, we have applied back-off and deleted interpolation model. In the backoff method, for *N* = 3 in (1), i.e. for a trigram model, the word sequences will follow trigram probabilities at first; if it could not match then word sequences will follow bigram model; if it also could not match then word sequence will follow unigram model and predict at least a word. As the system is for word prediction and it is playing with word sequence, so though any probability of word sequence is zero for multiplication, it will predict at least a word. The Backoff *N*-gram modelling is a nonlinear method. The difference is that in backoff , if there are non-zero trigram counts, it rely on trigram counts and don't interpolate bigram and unigram counts at all [14]. The *N*-gram version (*N* = *c*) of backoff model can be represented as follows:

$$P\left(w_i \mid w_{i-(c-1)}.......w_{i-1}w_{i-2}\right) = \begin{cases} P\left(w_i \mid w_{i-(c-1)}............w_{i-2}w_{i-1}\right), & \text{if } C\left(w_{i-(c-1)}............w_{i-2}w_{i-1}w_i\right) > 0 \\ \alpha_1 P\left(w_{i-(c-2)}............w_{i-2}w_{i-1}\right), & \text{if } C\left(w_{i-(c-1)}............w_{i-2}w_{i-1}w_i\right) = 0 \\ & \text{and } C\left(w_{i-(c-2)}............w_{i-2}w_{i-1}w_i\right) > 0 \\ \vdots \\ \alpha_{c-1} P(w_i), & \text{otherwise} \end{cases} \quad (2)$$

The deleted interpolation algorithm combining different *N*-gram orders by linearly interpolating all three models when they are computing any trigram. The detailed interpolation formula is as follows:

$$\hat{P}\left(w_n \mid w_{n-2}w_{n-1}\right) = \lambda_1\left(w_{n-2}^{n-1}\right)P\left(w_{n-2}w_{n-1}\right) + \lambda_2\left(w_{n-2}^{n-1}\right)P\left(w_n \mid w_{n-1}\right) + \lambda_3\left(w_{n-2}^{n-1}\right)P(w_n) \quad (3)$$

### 3.1. WHY *N*-GRAM BASED WORD PREDICTION?

We choose *N*-gram based word prediction system because these are more statistical approach to predict upcoming word in a sentence with more accuracy and *N*-gram language models provide a natural approach to the construction of sentence completion systems. To measure the probabilities by statistical model like *N*-gram data is divided into training set and test set. We compute the probabilities of a test sentence from trained corpus which is designed very carefully. Suppose a word "সহকারী" occurs 400 times in a corpus of one million Bangla words. Then the estimated probability for the word "সহকারী" is $\frac{400}{1000000} = 0.0004$.

## 4. METHODOLOGY

Our approach starts with a sentence fragment and come up with a word using a stochastic language model as shown in Fig. 2. We use five stochastic language models, namely unigram, bigram, trigram, backoff and deleted interpolation.

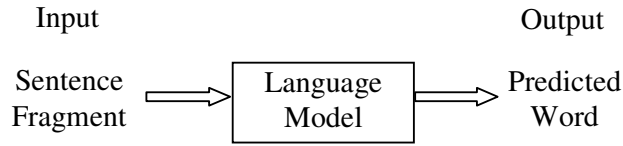

Fig. 2. Word prediction as the task of mapping a sentence fragment into a predicted word.

Conditional frequency distributions come into play when language model learning takes place for word prediction. A conditional frequency distribution refers to collection of frequency distribution for the same experiment, run under different conditions. We intend to predict a word's text (outcome), based on the text of the word that it follows (context). To predict the outcomes of an experiment, a training corpus is examined first, where the context and outcome for each run of the experiment are known. When presented a new run of the experiment, the outcome that occurred most frequently for the experiment's context is simply chosen. The total process is shown in Fig. 3.

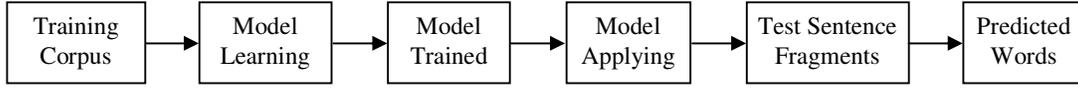

Fig. 3. The approach for building our language model for predicting word.

## 5. IMPLEMENTATION

Our work starts with a training corpus of size 0.25 million words. The corpus has been constructed from the popular Bangla newspaper the daily "Prothom Alo". The corpus contains 14,872 word forms. We have taken some test data in a file from the training corpus. After that comprehensive unigram, bi-gram and tri-gram statistics have been automatically generated and stored the sentences with predicted words in a file. Thus we have constructed *N*-gram models of word prediction by counting frequencies of words in a very large corpus, i.e. database and determine probabilities using *N*-gram. We employ a Java program in order to perform our experiments using unigram, bigram, trigram, backoff and deleted interpolation language models.

First, we start working on unigram model. We split the entire corpus into two parts, namely training part and testing part. We use holdout method [14] for selecting the proportion of data reserved for training and for testing. We split the corpus at the proportion of two-thirds for training and one-third for testing. In order to avoid model overfitting problem, i.e. to have low training error as well as low generalization error, we use a validation set. In accordance with this approach, we divide the original training data into two smaller subsets. One of the subsets is used for training, while the other one, i.e. the validation set, is used for calculating the generalization error. We fix two-thirds of the training set for model building while the remaining one-third is used for error estimation. We repeat holdout method five times in order to find the best model. After finding out our best model, we compute the accuracy of the model from the test set. We evaluate the performance of the classifier by varying the length of test sentence. Likewise we work on bigram, trigram, backoff and deleted interpolation models. All results are shown in Table I and Fig. 4. For deleted interpolation model, we empirically found the linear weights $\lambda_1 = 0.5$, $\lambda_2 = 0.33$ and $\lambda_3 = 0.17$ such that they sum to 1:

$$\sum_{i=1}^{3} \lambda_i = 1. \qquad (4)$$

So, the probability equation for our deleted interpolation becomes as follows:

$$\hat{P}(w_n | w_{n-1} w_{n-2}) = 0.5 P(w_n | w_{n-1} w_{n-2}) + 0.33 P(w_n | w_{n-1}) + 0.17 P(w_n). \qquad (5)$$

We get the following probability for the following sentence or sample test data.

Test sample no. 1

Test sentence fragment: বেনাপোল বন্দরের পরীক্ষণ বিভাগের দায়িত্বপ্রাপ্ত সহকারী

Word to be predicted: কমিশনার

For unigram model, we get the following results.

Word to predicted: ও

$P$(বেনাপোল বন্দরের পরীক্ষণ বিভাগের দায়িত্বপ্রাপ্ত সহকারী ও) = $P$(বেনাপোল) × $P$(বন্দরের) × $P$(পরীক্ষণ) × $P$(বিভাগের) × $P$(দায়িত্বপ্রাপ্ত) × $P$(সহকারী) × $P$(ও) = 5.09E-5 × 1.19E-4 × 8.48E-5 × 5.94E-4 × 1.02E-4 × 2.38E-4* 0.01

For bigram model, we get the following results.

Word predicted: জজ

$P($বেনাপোল বন্দরের পরীক্ষণ বিভাগের দায়িত্বপ্রাপ্ত সহকারী জজ$) = P($বেনাপোল $| <s>) \times P($বন্দরের $|$ বেনাপোল$) \times P($পরীক্ষণ $|$ বন্দরের$) \times P($বিভাগের $|$ পরীক্ষণ$) \times P($দায়িত্বপ্রাপ্ত $|$ বিভাগের$) \times P($সহকারী $|$ দায়িত্বপ্রাপ্ত$) \times P($জজ $|$ সহকারী$) = 0.0 \times 0.33 \times 0.14 \times 0.2 \times 0.03 \times 0.17 \times 0.14$

For trigram model, we get the following results.

Word predicted: কমিশনার

$P($বেনাপোল বন্দরের পরীক্ষণ বিভাগের দায়িত্বপ্রাপ্ত সহকারী কমিশনার$) = P($বেনাপোল $| <s> <s>) \times P($বন্দরের $| <s>$ বেনাপোল$) \times P($পরীক্ষণ $|$ বেনাপোল বন্দরের$) \times P($বিভাগের $|$ বন্দরের পরীক্ষণ$) \times P($দায়িত্বপ্রাপ্ত $|$ পরীক্ষণ বিভাগের$) \times P($সহকারী $|$ বিভাগের দায়িত্বপ্রাপ্ত$) \times P($কমিশনার $|$ দায়িত্বপ্রাপ্ত সহকারী$) = 0.0 \times 0.0 \times 1.0 \times 1.0 \times 1.0 \times 1.0 \times 1.0$

For backoff model, we get the following results.

Word predicted: কমিশনার

$P($বেনাপোল বন্দরের পরীক্ষণ বিভাগের দায়িত্বপ্রাপ্ত সহকারী কমিশনার$) = P($বেনাপোল $| <s> <s>) \times P($বন্দরের $| <s>$ বেনাপোল$) \times P($পরীক্ষণ $|$ বেনাপোল বন্দরের$) \times P($বিভাগের $|$ বন্দরের পরীক্ষণ$) \times P($দায়িত্বপ্রাপ্ত $|$ পরীক্ষণ বিভাগের$) \times P($সহকারী $|$ বিভাগের দায়িত্বপ্রাপ্ত$) \times P($কমিশনার $|$ দায়িত্বপ্রাপ্ত সহকারী$) = 0.0 \times 0.0 \times 1.0 \times 1.0 \times 1.0 \times 1.0 \times 1.0$

For deleted interpolation model, we get the following results.

Word predicted: কমিশনার

$P($কমিশনার $|$ দায়িত্বপ্রাপ্ত সহকারী$) = 0.5 \times P($কমিশনার $|$ দায়িত্বপ্রাপ্ত সহকারী$) + 0.33 \times P($কমিশনার $|$ সহকারী$) + 0.17 \times P($কমিশনার$) = 0.5 \times 1.0 + 0.33 \times 0.07 + 0.17 \times 0.1\text{E-}3$

Test sample no. 2
Test sentence fragment: দেশের বৃহত্তম
Word to be predicted: পর্বতমালা
For unigram model, we get the following results.
Word predicted: ও
$P($দেশের বৃহত্তম ও$) = P($দেশের$) \times P($বৃহত্তম$) \times P($ও$) = 0.0 \times 1.02 \text{ E-}4 \times 0.01$
For bigram model, we get the following results.
Word predicted: পর্বতমালা
$P($দেশের বৃহত্তম পর্বতমালা$) = P($দেশের $| <s>) \times P($বৃহত্তম $|$ দেশের$) \times P($পর্বতমালা $|$ বৃহত্তম$) = 0.0 \times 0.0 \times 0.33$

For trigram model, we get the following results.
Word predicted: Empty string, because there is no word following দেশের বৃহত্তম in the training corpus
For backoff model, we get the following results.
Word predicted: পর্বতমালা
$P($দেশের বৃহত্তম পর্বতমালা$) = P($দেশের $| <s>) \times P($বৃহত্তম $|$ দেশের$) \times P($পর্বতমালা $|$ বৃহত্তম$) = 0.0 \times 0.0 \times 0.33$

For deleted interpolation model, we get the following results.

Word predicted: পর্বতমালা

$P(পর্বতমালা | দেশের\ বৃহত্তম) = 0.5 \times P(পর্বতমালা | দেশের\ বৃহত্তম) + 0.33 \times P(পর্বতমালা | বৃহত্তম) + 0.17 \times P(পর্বতমালা)$

$= 0.5 \times 0.0 + 0.33 \times 0.33 + 0.17 \times 5.1\text{E-}5$

Table I. Results of all models used

| No of Words in Test Sentence | Accuracy | | | | |
|---|---|---|---|---|---|
| | Unigram | Bigram | Trigram | Backoff | Deleted Interpolation |
| 5 | 21.89% | 46.89% | 52.16% | 52.16% | 52.16% |
| 6 | 26.53% | 49.53% | 62.68% | 63.18% | 62.68% |
| 7 | 23.89% | 46.89% | 65.32% | 66.32% | 67.95% |
| 8 | 17.26% | 44.26% | 60.05% | 61.05% | 62.68% |
| 9 | 28.53% | 49.53% | 67.95% | 68.45% | 67.95% |
| 10 | 26.53% | 49.53% | 69.58% | 71.58% | 70.58% |
| 11 | 20.26% | 44.26% | 65.32% | 65.32% | 62.68% |
| 12 | 28.16% | 52.16% | 67.45% | 68.95% | 67.95% |
| 13 | 21.26% | 44.26% | 62.68% | 63.18% | 62.68% |
| 14 | 23.26% | 44.26% | 67.95% | 68.95% | 67.95% |
| 15 | 18.63% | 41.63% | 67.95% | 67.95% | 67.95% |
| 16 | 18.63% | 41.63% | 60.05% | 60.55% | 60.05% |
| 17 | 17.26% | 44.26% | 65.32% | 65.32% | 62.68% |
| 19 | 21.89% | 46.89% | 67.95% | 68.45% | 65.32% |
| 20 | 18.63% | 41.63% | 41.63% | 41.63% | 41.63% |

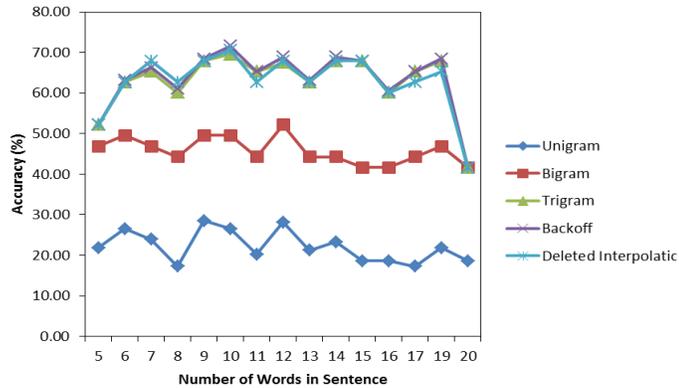

Fig. 4. Graphs of the results of all models used.

## 6. RESULT ANALYSIS

In order to understand the merits of our work in predicting words in Bangla, we need to delve into all results found. We see from Table I and Fig. 4 that trigram, backoff and deleted interpolation language model have performed almost in the same trend-line. Bigram model perform modestly, whereas unigram models performance is obviously very poor. The average accuracies of all models deployed are shown in Table II and Fig. 5. We see, from Fig. 5 and Table II, that the average accuracies of trigram, backoff and deleted interpolation model are close, where the accuracy (63.5%) of the backoff model is the maximum. Some results of predictions by all models deployed are given below.

Table II. Results of all models used

| Model | Average Accuracy |
|---|---|
| Unigram | 21.24 |
| Bigram | 45.84 |
| Trigram | 63.04 |
| Backoff | 63.50 |
| Deleted Interpolation | 62.86 |

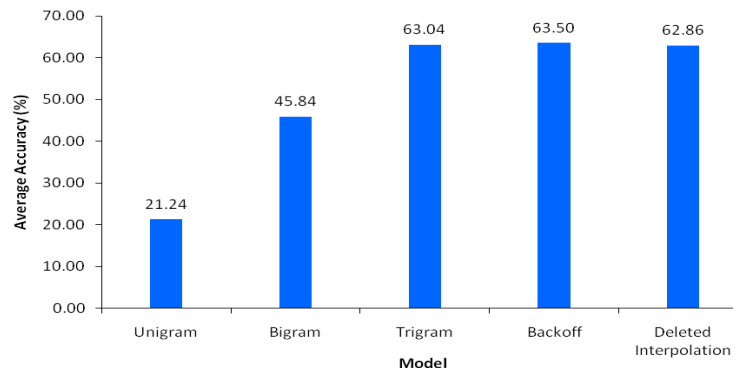

Fig. 5. Graphs of the results of all models used

Performance of word predictor depends on some important component such as which language model is used, the average length of the sentence in the language, amount of trained data set etc . In our work we use 5 to 20 words in a sentence of 100 sentence for each execution from .25 million trained data set. We see from Fig. 5 that initially performance of word prediction increases with the increase of *N* in *N*-gram language model.

## 4. CONCLUSIONS

*N*-gram based word prediction works well for English but we use for Bangla language which is more challenging to get 100% performance because it depends on training corpus of large data. We use more smoothed corpus data and increase data corpus size in future to get higher performance. Here in our work we use single word prediction but we can develop our process to predict a set of word to complete a sentence in a very meaningful way.

International Journal in Foundations of Computer Science & Technology (IJFCST) Vol.5, No.6, November 2015

## AUTHORS

**Md. Masudul Haque**  has obtained graduate degree in Computer Science  and Engineering from North South University  and  expert  level professional degree in  ICT from BUET. Now he is employing as a Programmer in the Local Government Engineering Department under the ministry of  Local Government of  Bangladesh. He has several publications in international and national journals and conference proceedings. His research interest is in Database Security, Web Security, Artificial Intelligence, Pattern Recognition and Natural Language Processing.

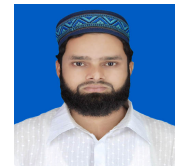

**Md. Tarek Habib** obtained his M.S. degree in Computer Science and Engineering (Major in Intelligent Systems Engineering) and B.Sc. degree in Computer Science from North South University in 2009 and BRAC University in 2006, respectively. Now he is an Assistant Professor at the Department of Computer Science and Engineering in Daffodil International University. He is much fond of research. He has had a number of publications in international and national journals and conference proceedings. His research interest is in Artificial Intelligence, especially Artificial Neural Networks, Pattern Recognition, Computer Vision and Natural Language Processing.

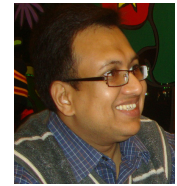

**Md. Mokhlesur Rahman**  was born at Satkhira in Bangladesh. He received his B.Sc degree in    Computer Science & Engineering from Mawlana Bhashani Science and Technology University, Bangladesh in 2010. His research interest is in Artificial Intelligence.He worked as a SENIOR WEB DEVELOPER at Squadro Solutions, Dhaka, Bangladesh in 2010. In the following year, he joined to International Acumen Limited as an ASSISTANT ENGINEER, Dhaka, Bangladesh. At the end of 2012, he joined as a LECTURER to Prime University, Mirpur-1, Dhaka, Bangladesh.

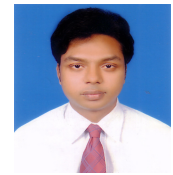